          [2001/04/25 1.1 (PWD)]
\documentclass[a4paper,twocolumn]{esapub2005} 
\usepackage{times}            
\usepackage{graphicx}         
\usepackage{amsmath}          
\usepackage{amssymb}
\usepackage{url}
\usepackage{svg}

\usepackage[backend=biber,
            style=numeric,
            sorting=none,
            giveninits=true,
            doi=false,
            isbn=false,
            url=false,
            eprint=false,
            maxnames=99,
            minnames=1]{biblatex}

\DeclareNameAlias{default}{family-given}

\DeclareFieldFormat[article]{title}{#1}
\DeclareFieldFormat[inproceedings]{title}{#1}
\DeclareFieldFormat[incollection]{title}{#1}
\DeclareFieldFormat[book]{title}{\mkbibemph{#1}}
\DeclareFieldFormat{journaltitle}{\mkbibemph{#1}}
\DeclareFieldFormat{booktitle}{\mkbibemph{#1}}

\DeclareFieldFormat{volume}{\textbf{#1}}

\DeclareFieldFormat{pages}{pp#1}

\DeclareFieldFormat{labelnumberwidth}{#1.}

\defbibenvironment{bibliography}
  {\list
     {\printfield[labelnumberwidth]{labelnumber}}
     {%
      \setlength{\labelwidth}{\labelnumberwidth}%
      \setlength{\leftmargin}{\dimexpr \labelnumberwidth+\labelsep+\bibhang\relax}%
      \setlength{\itemindent}{-\bibhang}%
      \setlength{\listparindent}{0pt}%
      \setlength{\itemsep}{\bibitemsep}%
      \setlength{\parsep}{\bibparsep}%
      }}
  {\endlist}
  {\item}

\renewbibmacro*{date+extrayear}{%
  \iffieldundef{year}
    {}
    {\printtext[parens]{\printdate}}}

\DeclareBibliographyDriver{article}{%
  \printnames{author}%
  \setunit{\space}%
  \usebibmacro{date+extrayear}%
  \newunit\newblock
  \printfield{title}%
  \newunit\newblock
  \printfield{journaltitle}%
  \setunit{\space}%
  \printfield{volume}%
  \printfield{number}%
  \newunit\newblock
  \printfield{pages}%
  \finentry}

\DeclareBibliographyDriver{book}{%
  \printnames{author}%
  \setunit{\space}%
  \usebibmacro{date+extrayear}%
  \newunit\newblock
  \printfield{title}%
  \newunit\newblock
  \printlist{publisher}%
  \newunit\newblock
  \printlist{location}%
  \newunit\newblock
  \printfield{pages}%
  \finentry}

\DeclareBibliographyDriver{inproceedings}{%
  \printnames{author}%
  \setunit{\space}%
  \usebibmacro{date+extrayear}%
  \newunit\newblock
  \printfield{title}%
  \newunit\newblock
  In \printfield{booktitle}%
  \newunit\newblock
  \printlist{publisher}%
  \newunit\newblock
  \printlist{location}%
  \newunit\newblock
  \printfield{pages}%
  \finentry}

\title{Towards an Adaptive Locomotion Strategy for Quadruped Rovers: Quantifying When to Slide or Walk on Planetary Slopes}
\author[1]{Alberto Sanchez-Delgado}
\author[1]{Jo\~ao Carlos Virgolino Soares}
\author[2]{David Omar Al Tawil}
\author[2]{Alessia Li Noce} 
\author[1]{Matteo~Villa} 
\author[1]{Victor Barasuol}
\author[2]{Paolo Arena}
\author[1]{Claudio Semini}
\affil[1]{Dynamic Legged Systems Lab, Istituto Italiano di Tecnologia (IIT), Via S. Quirico 19d, 16163 Genoa, Italy, alberto.sanchez@iit.it}
\affil[2]{Dipartimento di Ingegneria Elettrica Elettronica e Informatica, Universit\`a di Catania, Via Santa Sofia 64,
95123 Catania, Italy, david.altawil@phd.unict.it}

\begin{document}

\maketitle

\begin{abstract}
Legged rovers provide enhanced mobility compared to wheeled platforms, enabling navigation on steep and irregular planetary terrains. However, traditional legged locomotion might be energetically inefficient and potentially dangerous to the rover on loose and inclined surfaces, such as crater walls and cave slopes. This paper introduces a preliminary study that compares the Cost of Transport (CoT) of walking and torso-based sliding locomotion for quadruped robots across different slopes, friction conditions and speed levels. By identifying intersections between walking and sliding CoT curves, we aim to define threshold conditions that may trigger transitions between the two strategies. The methodology combines physics-based simulations in Isaac Sim with particle interaction validation in ANSYS-Rocky. Our results represent an initial step towards adaptive locomotion strategies for planetary legged rovers.
\end{abstract}

\section{Introduction}
Planetary exploration missions are increasingly targeting scientifically valuable regions such as crater walls, permanently shadowed regions (PSRs), and lava tubes, which often feature steep and irregular slopes. For instance, the Shackleton crater at the lunar south pole presents average wall inclinations of about $31^\circ$, with locations exceeding $35^\circ$, combined with loose regolith deposits that pose mobility challenges \cite{Wagner2013}. These environments are of high interest due to the potential presence of water ice and access to preserved geological records \cite{Mo2025}. However, their inaccessibility makes them some of the least explored and least understood locations. 

Traditional wheeled rovers, such as NASA's Mars rovers and the upcoming VIPER mission, provide proven reliability but face critical limitations on steep or granular terrains, where mobility rapidly decreases beyond $25$--$30^\circ$ slopes. Alternative platforms, such as tethered rovers (e.g., Axel) or hybrid systems (e.g., ATHLETE), have been proposed to overcome these limitations, yet they still struggle in highly irregular terrains \cite{Wilcox2007}. 

Legged robots have emerged as a promising alternative, combining versatile terrain adaptability with payload capacity \cite{Kolvenbach2018_scalability, Arm2023, Kolvenbach2020}. Their morphology enables negotiating irregular obstacles, dynamically redistributing contact forces, and even adopting novel locomotion modes beyond walking, such as crawling or sliding. This versatility makes them suitable for accessing crater rims, PSRs, and other planetary features beyond the reach of wheeled systems. At the same time, energy efficiency remains a fundamental constraint in planetary missions. On-board power is severely limited by mass and storage capacity, while solar energy generation is strongly reduced in shadowed regions or during long-duration traverses \cite{Litaker2025}. Consequently, optimizing locomotion strategies to reduce energy consumption is critical to extend mission lifetimes and maximize scientific return. 

Despite their mobility advantages, legged rovers face significant challenges when deployed in steep planetary terrains. Walking on slopes is energetically demanding and may lead to reduced stability due to slippage effects \cite{Eke2014}. Wheeled systems, while effective on moderate slopes, lose traction beyond $25$–$30^\circ$, and legged robots risk instability or failure on even steeper surfaces without specialized strategies \cite{Wilcox2007}. In such conditions, uncontrolled slippage can both waste energy and increase the risk of mission-ending accidents. Energy availability further constrains locomotion, requiring careful trade-offs between locomotion performance and energy consumption to ensure the viability of long-duration missions \cite{Litaker2025}.

One of the early initiatives in this field was the development of the six-legged SpaceClimber~\cite{Bartsch2012}, designed for lunar crater exploration. It demonstrated the potential of bio-inspired locomotion for negotiating steep slopes up to $40^\circ$ and operating in crater-like analog environments, with a strong emphasis on energy-efficient and adaptive mobility. Subsequent research highlighted that legged robots could be scaled and tailored for extraterrestrial environments, combining robust actuation with efficient control to traverse granular and irregular terrains \cite{Kolvenbach2018_scalability}. In this context, the work shown in \cite{dettmannGeneric2022} presented a generic guidance, navigation, and locomotion control system designed for both quadrupeds and hexapods, enabling robust terrain perception, path planning, and execution on rugged slopes and crater-like environments.

More recent efforts have focused on quadruped platforms. Studies on gait energetics in reduced gravity scenarios showed that dynamic gaits with flight phases can become advantageous on the Moon and Mars, particularly at moderate velocities \cite{Kolvenbach2018_gait}. In parallel, jumping locomotion has been investigated as a means of exploiting low gravity, with systems such as SpaceBok demonstrating the benefits of elastic energy storage and in-flight stabilization \cite{Kolvenbach2019_jumping}. Similar efforts explore reinforcement learning for in-flight attitude control, as in the Olympus quadruped, which targets Mars exploration scenarios \cite{Olsen2025}. Beyond locomotion studies, planetary analog deployments have validated the capability of legged robots in realistic mission-like conditions. The work in \cite{Arm2023} demonstrated a team of quadruped robots equipped with mapping pipelines, scientific instruments, and efficient locomotion controllers during the ESA Space Resources Challenge, showing their ability to traverse steep granular slopes and conduct scientific tasks. 

The studies reviewed in the previous paragraphs illustrate the progress and potential of legged systems for planetary exploration. However, most of these studies have focused on walking or jumping gaits, while alternative locomotion strategies remain less understood. Among them, sliding-based locomotion has recently been shown as a feasible locomotion modality for quadruped robots in analogue terrains \cite{barasuolControlled2024}. However, its energetic implications compared to waking was not studied yet. In this work, we address this gap by analyzing when sliding may represent an energetically favorable alternative to walking on steep surfaces. Specifically, we investigate how the CoT evolves as a function of slope inclination, velocity and terrain friction. By establishing preliminary criteria, we aim to contribute towards the definition of multi-modal locomotion strategies for future planetary missions, where energy efficiency and safe navigation over steep terrains are critical. The main contributions of this work are:
\begin{itemize}
    \item The extraction of CoT curves for walking and sliding under varying slopes, speeds, and friction conditions.
    \item The identification of slope-dependent intersection ranges between walking and sliding CoT curves as preliminary transition thresholds.
\end{itemize}

\section{System Description}\label{sec:system}

\begin{figure*}[ht]
    \centering
    \includegraphics[width=0.9\linewidth]{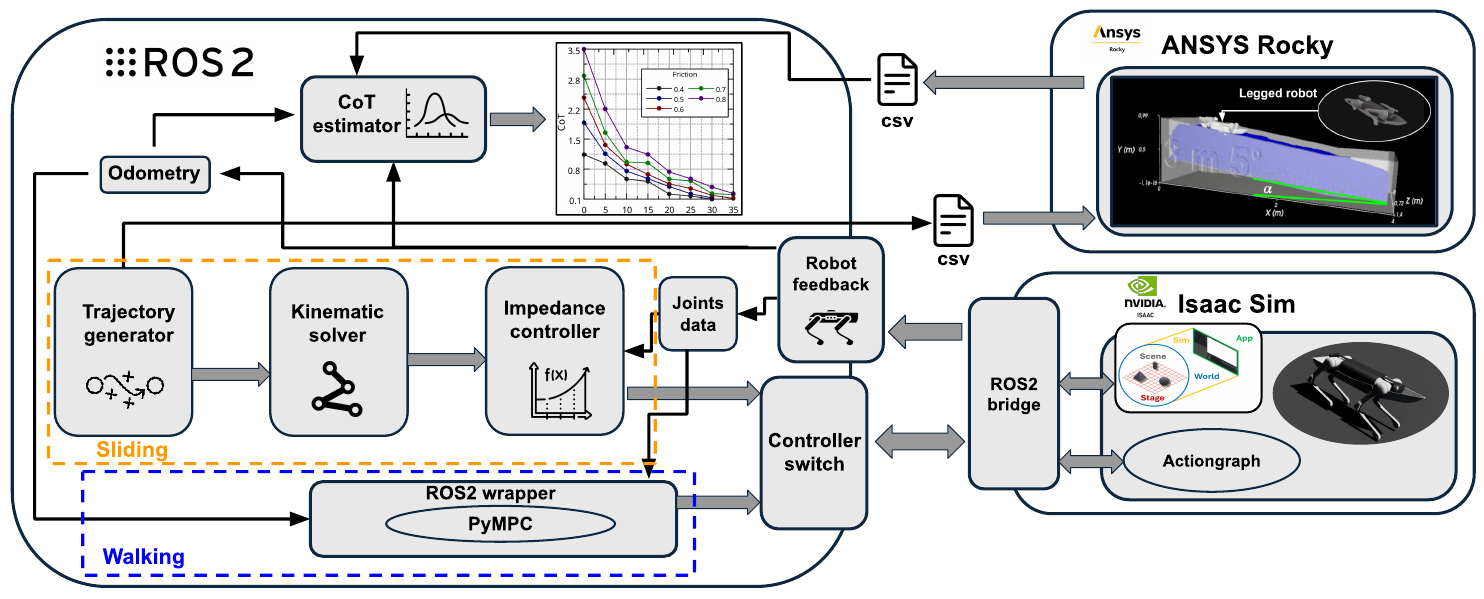}
    \caption{System structure integrating walking and sliding controllers with the simulation environment and data logging for CoT analysis.}
    \label{fig:system_diagram}
\end{figure*}

This section introduces our strategy to evaluate when either walking or sliding becomes energetically more advantageous for a quadruped rover. The system collects data from both locomotion modes and processes them into a common pipeline that estimates the CoT, thereby providing a physically grounded metric to compare with alternative strategies. An overview of the complete architecture is shown in Fig.~\ref{fig:system_diagram}. It integrates motion generation, simulation backend, and energy estimation modules through a ROS~2 interface. The system is composed of three main stages: \textbf{Motion generation}, where locomotion commands for walking or sliding are produced; \textbf{Simulation}, where these commands are executed in physics-based environments; and \textbf{CoT estimation}, where proprioceptive data are processed to compute energy-related metrics. Both locomotion strategies are embedded in a ROS~2-based architecture that interfaces with the simulators and logging modules, as illustrated in Fig.~\ref{fig:system_diagram}. This framework provides proprioceptive sensing (encoders, IMU, and foot contacts) and, additionally, logs joint torques and velocities for the CoT analysis. In addition, it connects to complementary particle-based simulations in ANSYS-Rocky, enabling consistency checks on granular interactions.

\subsection{Motion generation}

\begin{figure}[t]
    \centering
    \includegraphics[width=0.85\linewidth]{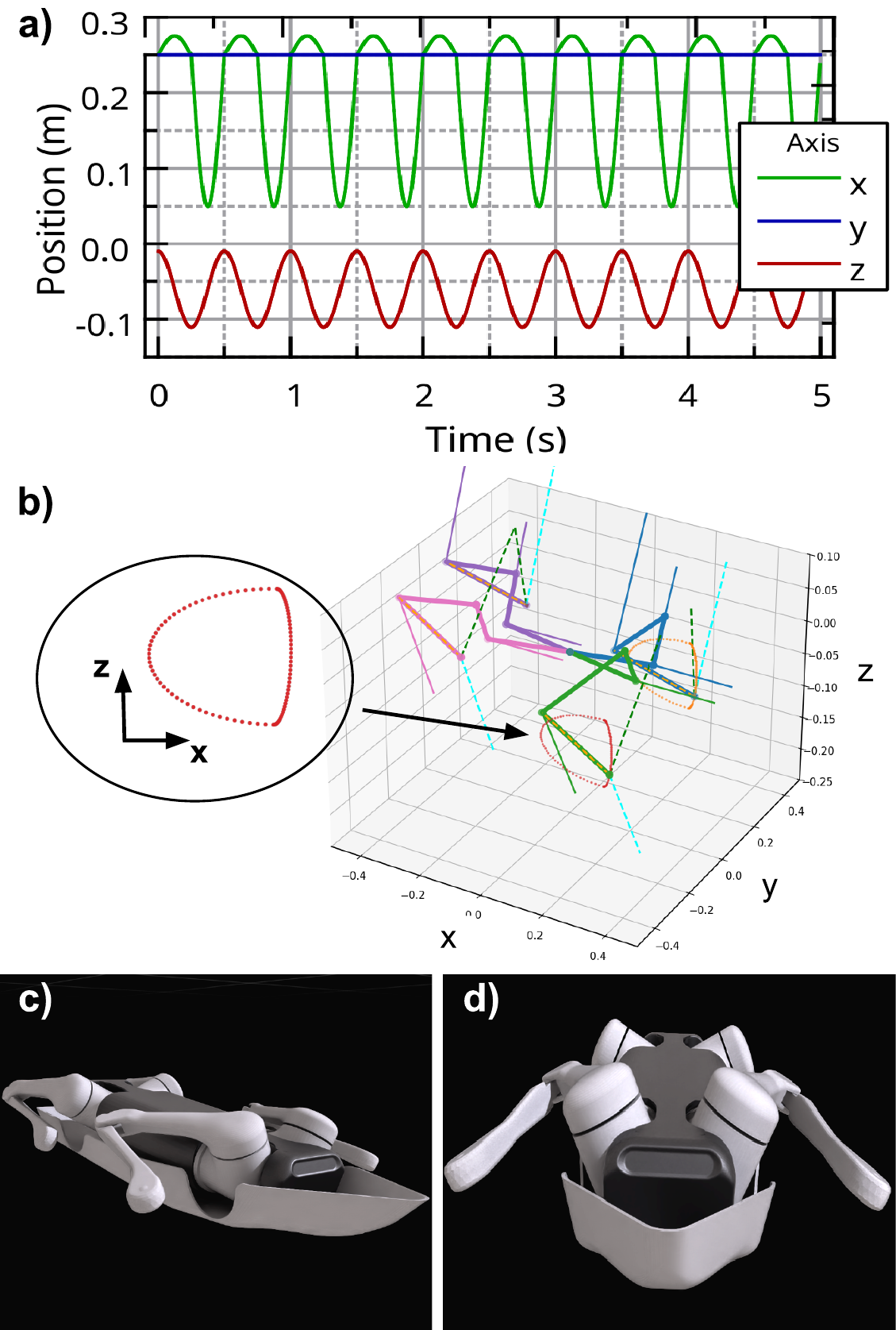}
    \caption{a) Generated trajectory for a single foot (Left Front - LF) in the task space.
    b) Isometric view of the trajectories for the robot legs. c) 3D view of the simulated robot with the legs in the home configuration. d) Frontal view of the robot with the legs in the home configuration.}
    \label{fig:tragjen}
\end{figure}

Walking is implemented using the Quadruped-PyMPC framework~\cite{turrisi2024sampling}, an open-source Python implementation of a Model Predictive Controller (MPC) for quadruped locomotion. 
It optimizes ground reaction forces online to track desired base velocities or poses while  considering the robot’s dynamics and contact constraints. This controller provides a consistent baseline for evaluating the energetic cost of walking under different slopes and speeds. Sliding locomotion relies on torso–ground and foot-ground contact while actively trusting using the robot's front legs. Three main modules are considered: trajectory generation, kinematic solver, and impedance controller.

\paragraph{Trajectory generator.}
For sliding-based locomotion, the trajectory generator produces Cartesian foot trajectories, as defined in Eqs.~\ref{eq:x_swing_front}--\ref{eq:home_filter}. Figure~\ref{fig:tragjen}a shows the trajectory of the left front leg along the $x$-, $y$-, and $z$-axes, while Fig.~\ref{fig:tragjen}b illustrates the complete trajectory formed by the front legs in 3D space. These motions originate from a nominal \emph{home configuration of the leg joint positions} (Fig.~\ref{fig:tragjen}c--d), where the hind legs remain retracted and the front legs are positioned with sufficient clearance from the ground to avoid unintended contact that could interfere with torso sliding. In this posture, only the left front (LF) and right front (RF) legs are actively commanded, while the hind legs remain fixed.

Let the sampling instants be $t_k = k \Delta t$ with $\Delta t = 1/f_s$ (task servo rate), and define $\phi_k = 2\pi f\, t_k$, $s_k = \sin(\phi_k)$, and $c_k = \cos(\phi_k)$, where $f$ is the commanded step frequency. For the front legs, the swing component in the longitudinal direction is piecewise scaled by the sign of the sine:
\begin{equation}
    x_{\mathrm{LF,RF}}[k] 
    = v \times 
    \begin{cases}
        L^{+}\, s_k, & s_k \ge 0,\\[2pt]
        L_{\mathrm{swing}}\, s_k, & s_k < 0,
    \end{cases}
    \label{eq:x_swing_front}
\end{equation}
where $L_{\mathrm{swing}}$ is the nominal swing length and $L^{+}$ is a reduced forward amplitude for the positive half-cycle and $v$ denotes the forward velocity. The lateral component is zero:
\begin{equation}
    y_{\mathrm{LF,RF}}[k] = 0.
    \label{eq:y_swing_front}
\end{equation}

When advancing forward ($v>0$), the vertical motion is generated around a fixed reference height via a constant offset plus a cosine modulation:
\begin{equation}
    z_{\mathrm{LF,RF}}[k] = -\,z_{0} \;+\; v\, H_{\mathrm{swing}}\, c_k,
    \label{eq:z_swing_front}
\end{equation}
with $z_{0}>0$ the constant height reference and $H_{\mathrm{swing}}$ the swing-height parameter. 

The complete Cartesian foot reference combines the nominal home posture, the swing component, and an optional steering correction ($\mathbf{p}_{\text{steer}}[k]$). The steering component is not considered in this preliminary study and is therefore not further explained:
\begin{equation}
    \mathbf{p}_{\text{foot}}[k] 
    = \mathbf{p_{\text{home}}}[k] 
    + \begin{bmatrix} x[k] \\[2pt] y[k] \\[2pt] z[k] \end{bmatrix}
    + \mathbf{p}_{\text{steer}}[k].
    \label{eq:pf_total}
\end{equation}

To avoid abrupt updates of the home posture, a first-order smoothing is applied:
\begin{equation}
    \mathbf{p}_{\text{home}}[k+1] 
    = (1-\alpha)\,\mathbf{p}_{\text{home}}[k] + \alpha\, \mathbf{p}^{\text{inp}}_{\text{home}}[k]
    \label{eq:home_filter}
\end{equation}
where $\mathbf{p}^{\text{inp}}_{\text{home}}$ denotes the home configuration and $\alpha \in (0,1)$.

\paragraph{Kinematic solver.}
The Cartesian foot trajectories generated in the previous section, $\mathbf{p}_{\text{foot},l}[k]$, are converted into joint references through inverse kinematics.  For each leg $l$, the position mapping is expressed as in Eq.~\ref{eq:ik}.
\begin{equation}
    \boldsymbol{q_l}[k] = f^{(l)}_{\mathrm{IK}}\!\left(\boldsymbol{p}_{\text{foot},l}[k]\right),
    \label{eq:ik}
\end{equation}
where $\boldsymbol{q_l[k]}$ are the joint angles and $f^{(l)}_{\mathrm{IK}}(\cdot)$ denotes the closed-form solution derived from the leg geometry. The velocity mapping is expressed as in Eq.~\ref{eq:diff_ik}.

\begin{equation}
    \boldsymbol{\dot{q}}_l[k] = J_l^{-1} \boldsymbol{\dot{p}_{\text{foot},l}}[k],
    \label{eq:diff_ik}
\end{equation}

where $J_l$ is the Jacobian of leg $l$.

\paragraph{Impedance controller.}
The joint references $\mathbf{q}_\ell^{\mathrm{des}}[k]$ obtained from inverse kinematics are tracked with a joint-space impedance controller, 
which directly outputs torque commands $\boldsymbol{{\tau}_l[k]}$ for each actuator per leg. The control law is expressed as in Eq.~\ref{eq:impedance_joint}

\begin{equation}
    \boldsymbol{\tau}_l[k]
    = K_q \big(\mathbf{q}_l[k] - \mathbf{q}[k]\big)
    + D_q \big(\dot{\mathbf{q}}_l[k] - \dot{\mathbf{q}}[k]\big)
    \label{eq:impedance_joint}
\end{equation}

where $\mathbf{q}[k]$ and $\dot{\mathbf{q}}[k]$ are the measured joint positions and velocities, 
$K_q$ and $D_q$ are diagonal stiffness and damping matrices.

\paragraph{Controller switch.}
A dedicated block manages the selection between the walking and sliding controllers. Depending on the user-defined condition, it routes either the MPC (walking) or the impedance controller (sliding) outputs to the robot joints in Isaac~Sim. This ensures that only one set of torque commands is active at a time, enabling controlled testing of both locomotion modes. 

\subsection{Simulation}
The control commands generated by the MPC (walking) or the trajectory generator and impedance controller (sliding) are executed in physics-based simulators that provide proprioceptive feedback. Isaac Sim supports both walking and sliding on rigid ramps under closed-loop control, while ANSYS-Rocky, based on the Discrete Element Method (DEM), is employed exclusively for sliding to capture particle-level effects such as sinkage and resistance. Reaction forces and velocities of the feet along with the pose of the body are recorded as a dataset which is sent back in the form of a CSV file to be processed by the CoT estimator.

\subsection{CoT estimation}
\label{sec.COTestimation}
The proprioceptive data collected from the simulators, namely joint torques and angular velocities in addition to the odometry, are processed to estimate the energetic cost of locomotion. Specifically, the instantaneous mechanical power at each joint is computed as the product of torque and angular velocity, and integrated over time to obtain the total mechanical energy expenditure. The mechanical energy $E$ is estimated from joint-level signals as follows: 

\begin{equation}
    E = \int_{t_0}^{t_f} \sum_{i=1}^{N} \big| \boldsymbol{\tau_i(t)}\,\boldsymbol{\dot{q}_i(t)} \big| \, dt,
    \label{eq:energy}
\end{equation}

where $\tau_i(t)$ and $\dot{q}_i(t)$ are the torque and angular velocity of joint $i$, $N$ is the number of actuated joints, and $t_0$ and $t_f$ correspond to the initial and final times of the period where the locomotion (walking or sliding) is active, respectively. The absolute value ensures that all power contributions are considered positive, regardless of direction of motion. The CoT is then defined as the dimensionless ratio between energy consumption and the product of robot weight and distance traveled:

\begin{equation}
    \mathrm{CoT} = \frac{E}{mgd},
    \label{eq:cot}
\end{equation}

where $m$ is the robot mass, $g$ the gravitational acceleration, and $d$ the distance traveled. When Cartesian quantities are available instead of joint-level signals, as in  ANSYS-Rocky simulations, joint torques and angular velocities can be reconstructed  from foot forces and velocities using the Jacobian mapping. The torque vector $\boldsymbol{\tau}$ is obtained as shown in Eq.~\eqref{eq:tau_from_grf}, and the joint velocities as in Eq.~\eqref{eq:qdot_from_v}:

\begin{equation}
    \boldsymbol{\tau} = J_l^{\top}\,\mathbf{F}_f + \boldsymbol{\tau}_{\mathrm{free}},
    \label{eq:tau_from_grf}
\end{equation}

\begin{equation}
    \dot{\boldsymbol{q}} = J_l^{-1}\,\dot{\mathbf{x}}_f,
    \label{eq:qdot_from_v}
\end{equation}

where $\mathbf{F}_f$ is the ground reaction force at the foot, $\mathbf{\tau_{\mathrm{free}}}$ denotes the joint torques required to accelerate the limb, $\dot{\mathbf{x}}_f$ the Cartesian foot velocity, $J_l$ the leg Jacobian.

\section{Simulation Setup and Evaluation}\label{sec:simulation_setup}\label{sec:evaluation} 

\subsection{Robotic System and Simulation Platforms}
The simulations were conducted using the quadruped robot Aliengo (Unitree Robotics), a 24\,kg platform with three actuated degrees of freedom per leg. To enable sliding locomotion on inclined terrains, the robot was equipped with a custom-designed lower torso previously introduced in~\cite{barasuolControlled2024}. This modification increases the effective contact area with the ground, reduces undesired sinkage, and improves the interaction with granular surfaces. In this study we considered the streamlined version without skids, focusing on the effects of contact area and surface friction during controlled sliding. It is important to emphasize that all the simulations were carried out under lunar gravity conditions ($1.625 \,\text{m/s}^2$).

The Isaac Sim simulator provides proprioceptive data streams equivalent to those available on a real quadruped, including joint states, foot contacts, and base pose. For the purpose of this study, only torques and angular velocities of the joints as well as total distance traveled by the body of the robot are used to compute the CoT, while additional signals are logged solely for monitoring and consistency checks. Two complementary simulation platforms were used (see Fig.~\ref{fig:simulation_envs}). NVIDIA Isaac Sim (version 4.2) provides a physics-based multi-body environment with native ROS 2 integration, allowing online evaluation of both locomotion strategies under closed-loop control. Walking and sliding controllers are executed in real time, as if deployed on the physical robot, enabling consistent comparisons between gaits. However, Isaac Sim represents the terrain as continuous rigid surfaces without explicit particle interactions, which limits the realism of sliding on granular media.

To address this limitation, ANSYS-Rocky was employed as second simulator to replicate sliding motion under granular terrain interactions. ANSYS-Rocky implements a DEM formulation that accounts for particle size, shape, and inter-particle friction, capturing sinkage and resistance effects absent in Isaac Sim. Since full closed-loop control is not feasible in ANSYS-Rocky, joint trajectories generated during Isaac Sim runs are exported and replayed offline. The simulator then provides forces and velocities of the feet, and base displacements, from which the CoT can be recalculated (see Sect. \ref{sec.COTestimation}). A single Rocky simulation requires approximately 24–48 hours of computation on a workstation equipped with an AMD Ryzen Threadripper PRO 7975WX 32-core CPU (4 GHz) and an NVIDIA RTX A6000 GPU with 49 GB of memory. While ANSYS-Rocky was used exclusively for sliding scenarios, Isaac Sim was used for both walking and sliding, providing a complementary picture of locomotion performance across environments.

\subsection{Evaluation Scenarios}
\label{sec:scenarios_method}

\begin{figure}[t]
  \centering
  \includegraphics[width=0.9\linewidth]{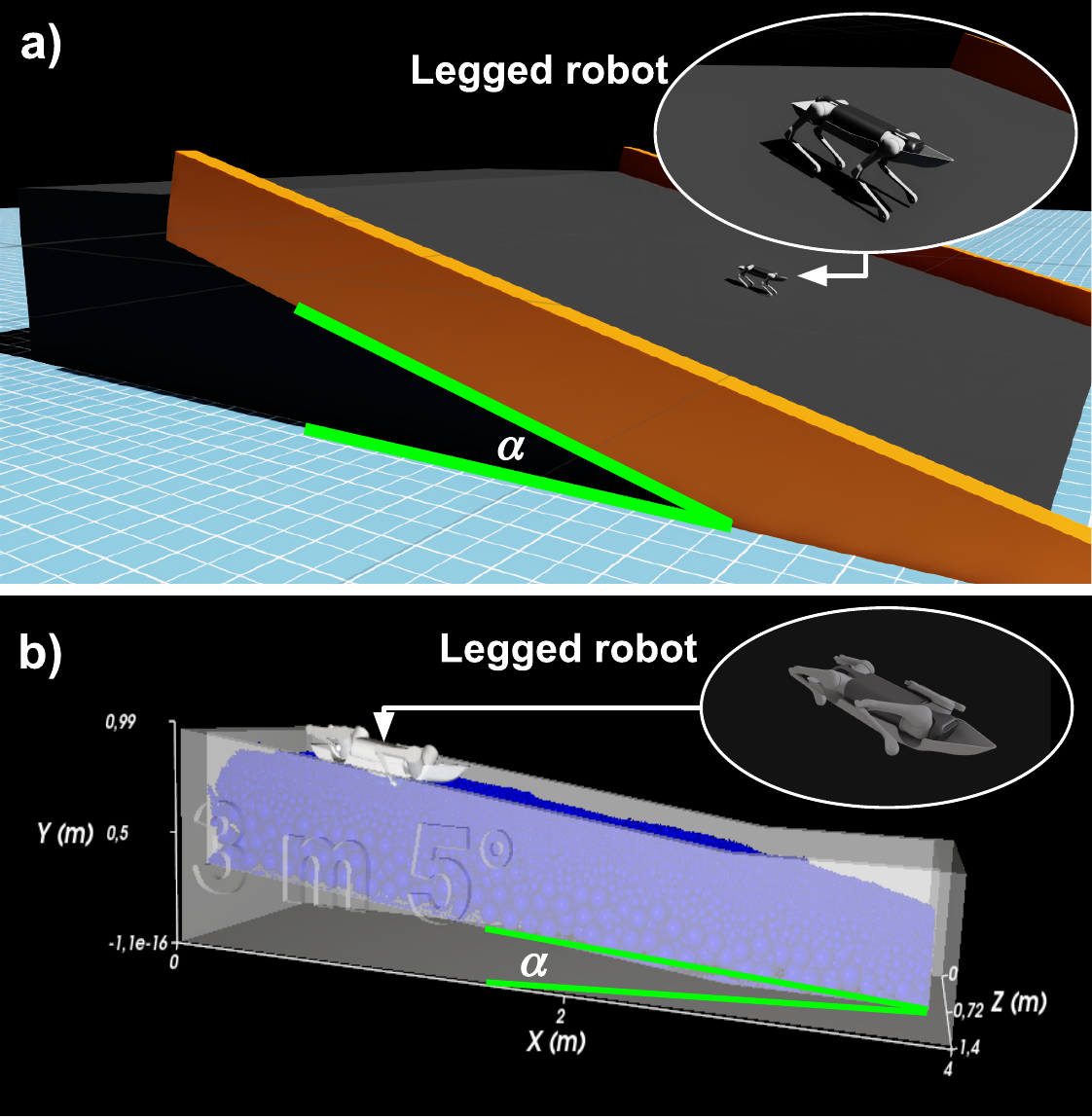}
  \caption{Simulation environments used for evaluation. 
  a) Isaac~Sim ramps with variable inclination $\alpha$, supporting both walking and sliding under closed-loop control. 
  b) ANSYS~Rocky granular ramps, used exclusively for sliding through offline replay of leg joint trajectories.}
  \label{fig:simulation_envs}
\end{figure}

This section describes the terrain setups, parameter variations, and simulation protocol employed to evaluate walking and sliding locomotion. Two complementary environments were considered: (i) rigid-surface ramps simulated in Isaac~Sim, where controllers run online in closed loop, and (ii) granular ramps in ANSYS~Rocky, where trajectories are replayed offline to capture particle-level effects. Fig.~\ref{fig:simulation_envs} illustrates both environments. 

\paragraph{Isaac~Sim ramp scenarios.}
Planar ramps with adjustable inclination $\alpha$ were created to test both walking and sliding under controlled rigid-surface conditions. 
For walking, $\alpha$ was varied from $0^{\circ}$ to $+35^{\circ}$ in increments of $5^{\circ}$, covering uphill and downhill cases. Three constant commanded speeds were considered at each slope: $v \in \{0.1, 0.2, 0.3\}\,\text{m/s}$.  For sliding, ramps with $\alpha \in [0^{\circ}, 35^{\circ}]$ were combined with different torso--ground friction coefficients  $\mu_s \in [0.4, 0.8]$ (step $0.1$), with dynamic friction set to $0.85\,\mu_s$.  Each condition was repeated 10 times to ensure statistical consistency.  Isaac~Sim allowed running both locomotion strategies online, with joint states, base motion, and foot contacts recorded.

\paragraph{ANSYS~Rocky granular scenarios.}
To account for the influence of particulate terrain, sliding was also tested in ANSYS~Rocky using a DEM formulation. Ramps with $\alpha \in \{5^{\circ},\,20^{\circ}\}$ were combined with torso–ground friction coefficients in $\mu_s = 0.6$. Unlike Isaac~Sim, ANSYS-Rocky cannot host closed-loop control of the robot. Instead, joint trajectories generated in Isaac~Sim were exported and replayed offline. This setup enabled computing torques, angular velocities, and base displacements under granular interactions, from which the CoT was obtained. Tab.~\ref{tab:ansys_parameters} summarizes the main fixed and variable parameters used in ANSYS-Rocky simulations. As in the Isaac Sim simulation, ten repetitions were performed for each configuration.

\paragraph{Simulation protocol.}
All simulations started from the same initial pose and ran for the same ramp length. Between repetitions, simulations were reset to avoid inter-trial memory effects. In Isaac~Sim, data were collected online under closed-loop control using ROS2 bags, reproducing realistic robot execution. In ANSYS-Rocky, the offline replay allowed capturing granular effects absent in rigid models, while ensuring consistency by using identical input trajectories. Together, both environments provide complementary perspectives: controlled repeatability with real-time controllers in Isaac~Sim, and physically grounded interaction with particulate media in ANSYS~Rocky.

\begin{table}[t]
  \centering
  \caption{Parameters used in ANSYS~Rocky simulations.\label{tab:ansys_parameters}}
  \begin{tabular}{p{0.7\linewidth}c}
    \hline
    Parameter & Value \\
    \hline
    \textbf{Gravel parameters} & \\
    Density (kg/m$^3$) & 2760 \\
    Young's modulus (GPa) & $29$ \\
    Poisson ratio & 0.2 \\
    Inter-particle static friction & 0.58 \\
    Inter-particle rolling resistance & 0.3 \\
    Inter-particle restitution coeff. & 0.46 \\
    Inter-particle dynamic friction & 0.58 \\
    \hline
    \textbf{Steel wall parameters} & \\
    Wall--particle static friction & 0.45 \\
    Wall--particle rolling resistance & 0.3 \\
    Wall--particle restitution coeff. & 0.6 \\
    Wall--particle dynamic friction & 0.51 \\
    \hline
  \end{tabular}
\end{table}


\section{Results}\label{sec:results} 

This section presents the evaluation of downhill sliding and walking locomotion on different ramps following the description provided in the previous section. The analysis focuses on the CoT as a function of slope inclination, surface friction, and commanded speed. 

\subsection{Walking locomotion: effect of slope and speed}

\begin{figure}[t]
    \centering
    \includegraphics[width=0.78\columnwidth]{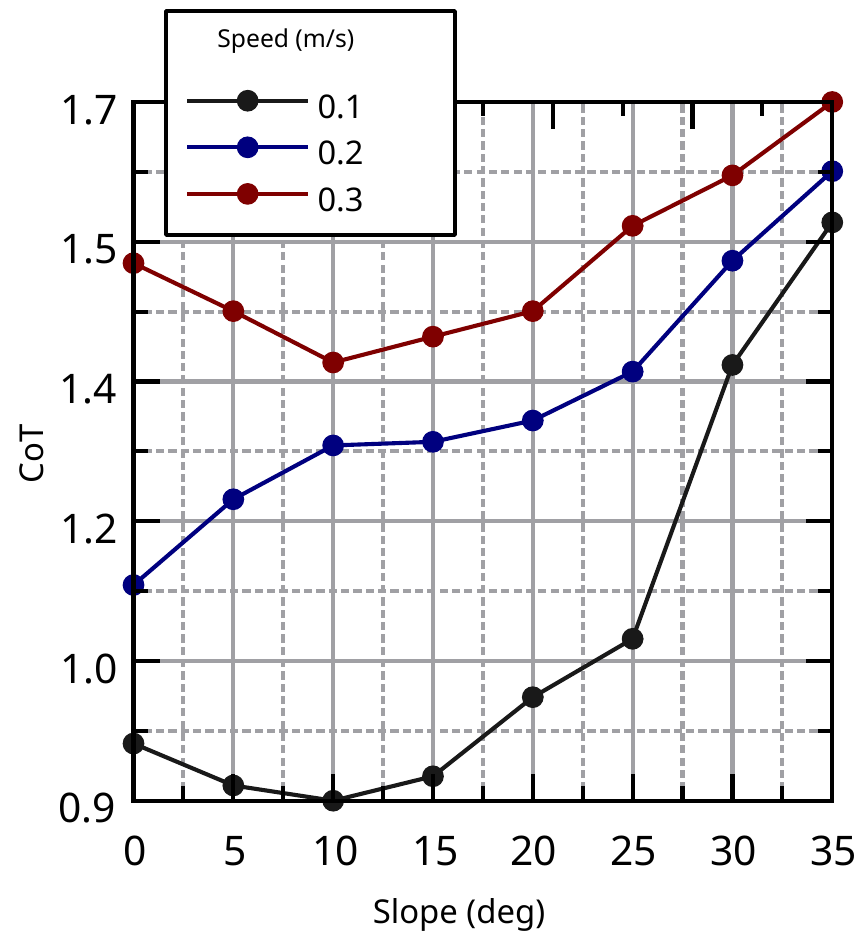}
    \caption{CoT of walking in Isaac Sim over downhill slopes from 0° to 35° and commanded base velocities of 0.1, 0.2, and 0.3 m/s.}
    \label{fig:cot_walking}
\end{figure}

Fig.~\ref{fig:cot_walking} shows the CoT of walking locomotion at commanded base velocities of 0.1, 0.2, and 0.3~m/s. At low speed (0.1~m/s), CoT remains low and nearly constant up to about $20^\circ$, followed by a steep increase beyond $25^\circ$. At higher speeds (0.2--0.3~m/s), CoT is consistently larger across all inclinations and grows gradually with slope angle. The lowest-speed case exhibits a shallow minimum around $10^\circ$, suggesting that a small downhill component can initially assist motion before energetic demands rise on steeper terrain. These results indicate a trade-off between speed and efficiency: while faster gaits reduce traversal time, they impose a larger energetic burden, especially on steep slopes.

\subsection{Sliding locomotion: effect of friction and slope}

\begin{figure}[t]
    \centering
    \includegraphics[width=0.78\columnwidth]{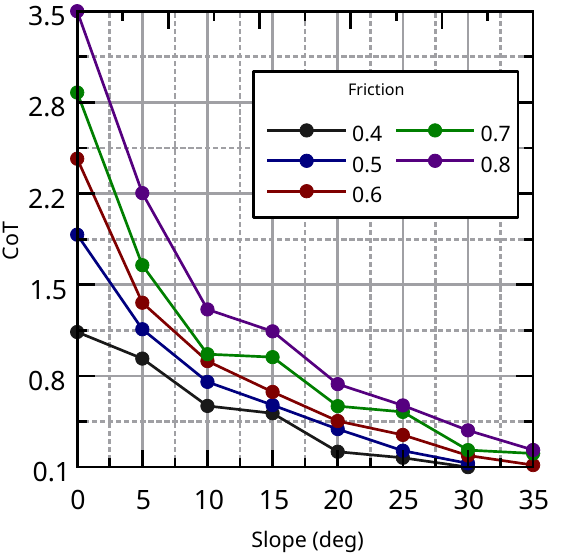}
    \caption{CoT for sliding in Isaac Sim across slope inclinations from 0° to 35° and static friction coefficients ($\mu_s = 0.4$--$0.8$).}
    \label{fig:cot_sliding}
\end{figure}

Fig.~\ref{fig:cot_sliding} reports the CoT of sliding locomotion under active thrust across different slope inclinations and static friction coefficients. The curves reveal a clear monotonic decrease in CoT with increasing slope angle. At shallow slopes ($<10^\circ$), sliding is energetically costly, particularly on high-friction surfaces ($\mu_s \geq 0.7$), where CoT values can exceed 2.5. As the slope angle increases, the gravitational component progressively assists motion, reducing the required thrust and lowering the CoT towards values close to 0.1 at $30$--$35^\circ$. Lower friction conditions ($\mu_s = 0.4$--$0.5$) reduce the energetic cost already at small inclinations, highlighting that controlled sliding can benefit from reduced torso--ground resistance. 

\subsection{Comparative analysis and intersections}

\begin{figure}[t]
    \centering
    \includegraphics[width=0.78\columnwidth]{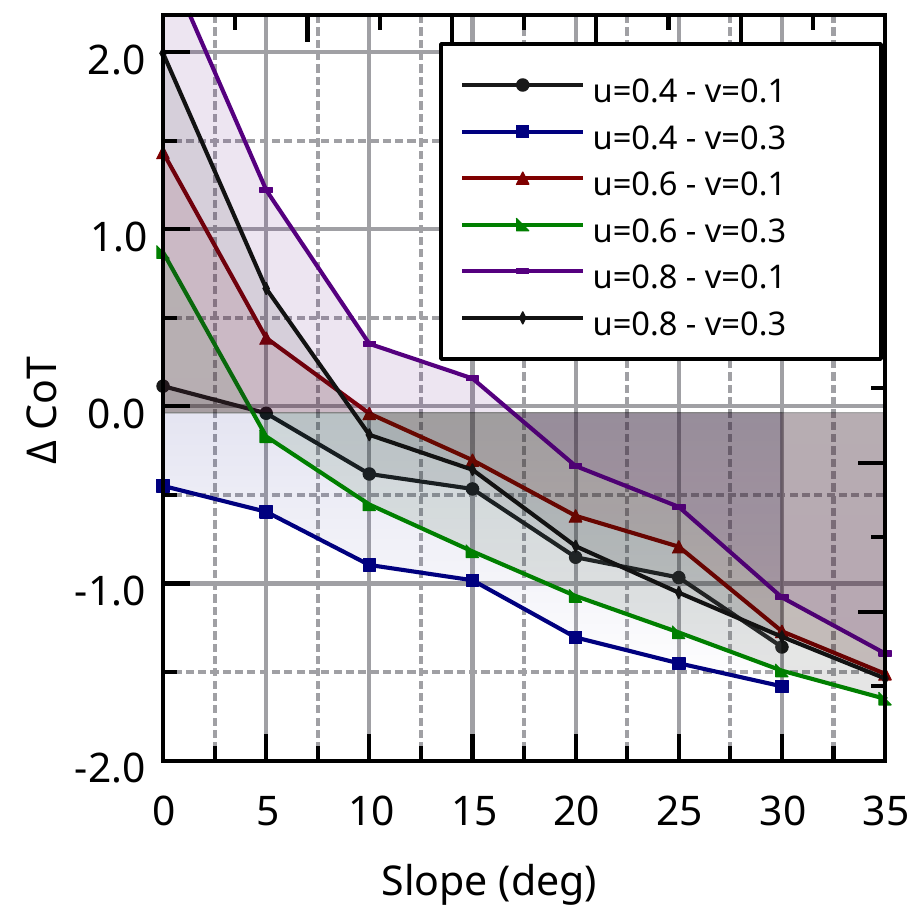}
    \caption{Comparison of walking and sliding illustrated with the 
    difference $\Delta CoT = CoT_{slide} - CoT_{walk}$ for 
    selected speeds and frictions.}
    \label{fig:cot_compare}
\end{figure}

Fig.~\ref{fig:cot_compare} reports the difference $\Delta \text{CoT} = \text{CoT}_{\text{slide}} - \text{CoT}_{\text{walk}}$ for representative combinations of walking speed ($v = 0.1, 0.3$\,m/s) and sliding friction ($\mu_s = 0.4, 0.6, 0.8$). Positive values indicate that sliding requires more energy than walking, while negative values indicate the opposite. The change of sign in $\Delta \text{CoT}$ therefore marks the transition angles $\alpha^{*}$ where both strategies present comparable energetic cost. The results show that for the intermediate friction case ($\mu_s = 0.6$), $\Delta \text{CoT}$ changes sign at slopes around $\alpha^{*} \in (4.0^\circ, 10.0^\circ$). Lower surface friction ($\mu_s = 0.4$) shifts the transition to smaller inclinations (earlier sliding advantage), while higher friction ($\mu_s = 0.8$) delays the crossover. Similarly, increasing walking speed from $0.1$ to $0.3$\,m/s anticipates the transition, since walking becomes rapidly more expensive. These observations highlight that transition thresholds are not fixed, but depend jointly on gait parameters and surface properties. For the extreme case $\mu_s = 0.4$ and $v = 0.3$ m/s, $\Delta$CoT remains strictly negative, meaning sliding is always more efficient. This outcome results from combining minimum friction, which lowers sliding cost, with maximum walking speed, which increases walking cost, and should be seen as a limit case of the tested conditions.

\subsection{Validation with granular terrain}

\begin{figure}[t]
    \centering
    \includegraphics[width=0.75\columnwidth]{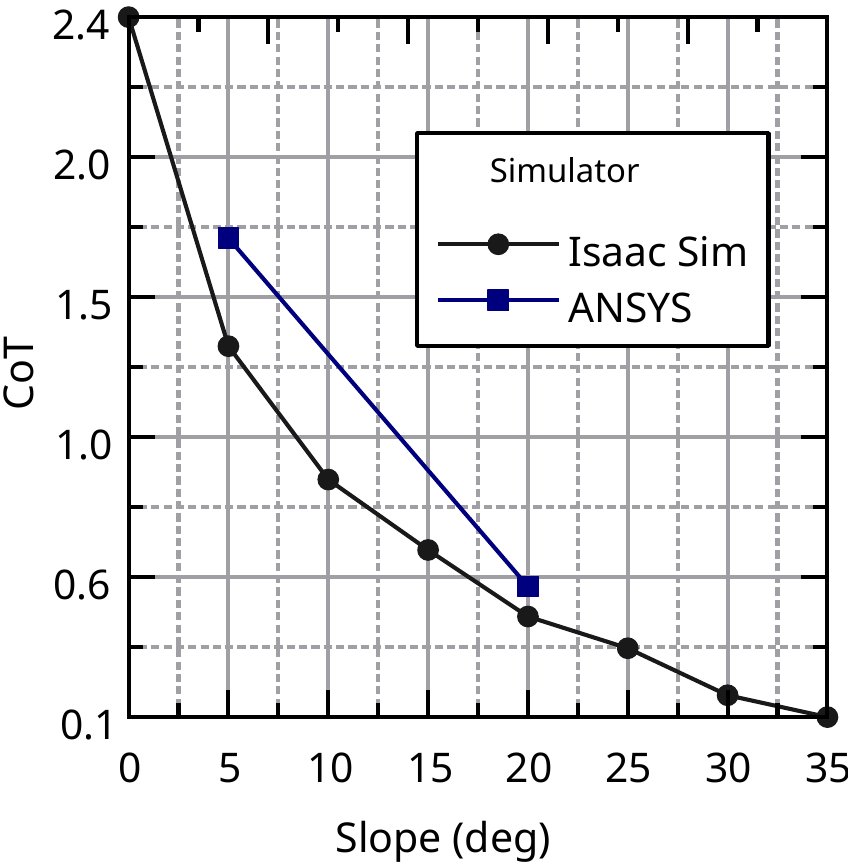}
    \caption{Comparison of sliding locomotion CoT between Isaac Sim and ANSYS Rocky}
    \label{fig:ansys_sim}
\end{figure}

To complement the rigid-surface results of Isaac Sim, preliminary simulations were conducted in ANSYS Rocky using a DEM formulation. Fig.~\ref{fig:ansys_sim} compares the CoT obtained in both simulators for sliding locomotion at a fixed friction coefficient $\mu_s=0.6$ and slope angles of $5^\circ$ and $20^\circ$. In both cases, ANSYS predicts systematically higher CoT values than Isaac Sim, reflecting the additional energetic penalties introduced by particle-level effects such as sinkage and local resistances. These points show an overall decreasing trend of CoT with slope and highlight that rigid-surface models may underestimate the energetic cost of sliding on granular terrain. 

\section{Discussions}\label{sec:discusions}
The results highlight a complementary energetic behavior between walking and sliding locomotion on inclined surfaces. Downhill walking tends to maintain relatively low CoT at shallow slopes, particularly at reduced commanded speeds, but becomes increasingly costly as inclination and velocity rise. Sliding, in contrast, appears penalized at gentle slopes due to torso–ground resistance, yet benefits from the gravitational component on steeper terrain, with CoT generally decreasing with slope angle. 

The comparison of both strategies suggests the presence of transition thresholds where one mode may become energetically more advantageous than the other. These thresholds, however, are not fixed values: they shift with walking speed and terrain friction, underscoring the need for multi-modal locomotion policies that take into account terrain conditions and gait parameters. Preliminary DEM-based simulations in ANSYS further indicated consistently higher CoT values than in Isaac Sim, suggesting that rigid-surface models may underestimate the energetic cost of sliding on granular terrain. The present study has a few constraints. It relies on a single locomotion controller and a specific quadruped platform with fixed mass distribution. For sliding, only one torso configuration was considered. All experiments were carried out under lunar gravity, and the particle-level validation with DEM-based simulations in ANSYS Rocky was limited to a reduced set of cases.


\section{Conclusions}\label{sec:conclusions} 
This work presented an energetic comparison between walking and sliding locomotion for quadruped rovers on inclined terrain. By extracting CoT curves across slope angles, friction levels, and walking speeds, we identified complementary trends and preliminary transition thresholds between the two strategies. Walking is efficient on gentle slopes and low speeds, while sliding becomes advantageous on steeper terrain. Preliminary DEM-based results in ANSYS also indicate that sliding on granular terrain may be more energetically costly than predicted by rigid-surface models.

Future work will focus on experimental validation in granular analog terrains, and on extending the analysis to different controllers, morphologies, torso designs, and multiple gravitational environments. These steps aim to consolidate predictive locomotion maps that support mode switching and energy-aware planning in challenging planetary scenarios.

\section*{Acknowledgments}
The authors would like to thank the Dynamic Legged Systems (DLS) Lab at the Istituto Italiano di Tecnologia for its continuous support. We also gratefully acknowledge the Dipartimento di Ingegneria Elettrica Elettronica e Informatica of the University of Catania for its collaboration and contribution to the development of this work. This work was partially financed by the Italian Space Agency (ASI).

\printbibliography

\end{document}